\documentclass[letterpaper, 10 pt, conference]{ieeeconf}  

\IEEEoverridecommandlockouts                              

\usepackage{graphicx} \graphicspath{ {./images/} } 
\usepackage[ruled,vlined]{algorithm2e}

\usepackage{amsmath} 
\usepackage{amssymb}  
\usepackage{siunitx}
\usepackage{url}
\usepackage{subcaption}
\usepackage[font=small,labelfont=bf]{caption}
\usepackage{xcolor}

\usepackage{tikz}
\usetikzlibrary{arrows, fit, positioning, decorations.pathreplacing, calc, math, shapes.misc, shapes.arrows}

\DeclareMathOperator*{\argmax}{arg\,max}

\title{\LARGE \bf
Automatic Acquisition of a Repertoire of Diverse Grasping Trajectories through Behavior Shaping and Novelty Search
}

\author{Aurélien Morel$^{1,*}$, Yakumo Kunimoto$^{2,*}$, Alex Coninx$^{3}$ and Stéphane Doncieux$^{4}$
\thanks{$^{*}$Both first authors contributed equally to this work.}%
\thanks{$^{1}$Aurelien Morel, Sorbonne Université, CNRS, Institut des Systèmes Intelligents et de Robotique, ISIR, F-75005 Paris, France and Ecole Polytechnique Fédérale de Lausanne, 1015 Lausanne, Suisse
        {\tt\small aurelien.morel.arthur@gmail.com}}%
\thanks{$^{2}$Yakumo Kunimoto, Sorbonne Université, CNRS, Institut des Systèmes Intelligents et de Robotique, ISIR, F-75005 Paris, France
        {\tt\small yakunimoto@yahoo.fr}}%
\thanks{$^{3}$Alex Coninx, Sorbonne Université, CNRS, Institut des Systèmes Intelligents et de Robotique, ISIR, F-75005 Paris, France
        {\tt\small coninx@isir.upmc.fr}}%
\thanks{$^{4}$Stéphane Doncieux, Sorbonne Université, CNRS, Institut des Systèmes Intelligents et de Robotique, ISIR, F-75005 Paris, France
        {\tt\small doncieux@isir.upmc.fr}}%
}

\begin{document}

\maketitle
\thispagestyle{empty}
\pagestyle{empty}

\begin{abstract}
Grasping a particular object may require a dedicated grasping movement that may also be specific to the robot end-effector. No generic and autonomous method does exist to generate these movements without making hypotheses on the robot or on the object. Learning methods could help to autonomously discover relevant grasping movements, but they face an important issue: grasping movements are so rare that a learning method based on exploration has little chance to ever observe an interesting movement, thus creating a bootstrap issue. We introduce an approach to generate diverse grasping movements in order to solve this problem. The movements are generated in simulation, for particular object positions. We test it on several simulated robots: Baxter, Pepper and a Kuka Iiwa arm. Although we show that generated movements actually work on a real Baxter robot, the aim is to use this method to create a large dataset to bootstrap deep learning methods.

\end{abstract}

\section{Introduction}

Grasping is a mandatory step for many daily object manipulation tasks. We do it without even thinking about it, but despite this apparent simplicity, it is a challenging movement that has been studied for decades and is still not completely solved for robotic agents, in particular when the 3D model of the target objects are not known \cite{billard2019trends}. If robots are to be deployed in our every-day environment, for instance as home assistants, they will have to deal with very diverse situations and will thus need a strong adaptivity. In this context, grasping synthesis algorithms that are robust and efficient on a large set of objects having diverse shape and weight will be required. Learning methods are expected to help improve the generalization ability of grasping controllers \cite{sahbani2012overview,bohg2013data,kleeberger2020survey}, but learning to grasp requires to face a major issue: grasping is a hard exploration problem, where only a small subset of a large policy space is relevant. It creates a challenge for exploration-based learning algorithms, as without strong constraints or prior knowledge on the task or on the policy space to explore, it is unlikely that a purely random exploration could discover relevant grasping motions. A learning algorithm may thus fail to bootstrap and find solutions to improve on, may it be through gradient descent or through a gradient-free trial and error process.  

\begin{figure}[ht]
    \centering
    
    \begin{subfigure}[b]{0.32\columnwidth}
        \centerline{\includegraphics[scale=0.34]{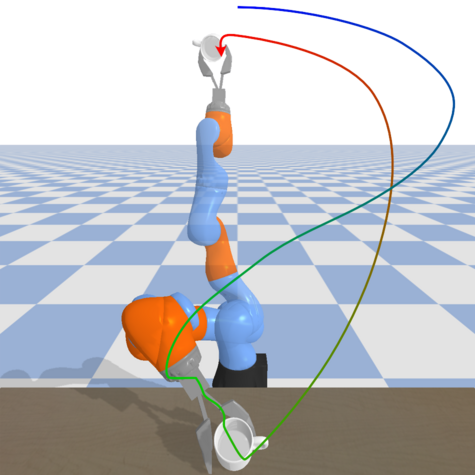}}
        \caption{Kuka}
        \label{kuka-mug}
    \end{subfigure}
    \begin{subfigure}[b]{0.32\columnwidth}
        \centerline{\includegraphics[scale=0.34]{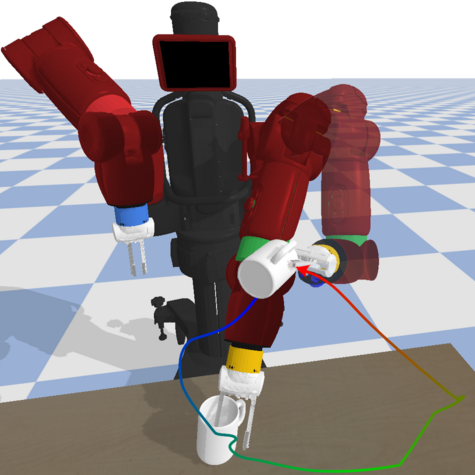}}
        \caption{Baxter}
        \label{baxter-mug}
    \end{subfigure}
    \begin{subfigure}[b]{0.32\columnwidth}
        \centerline{\includegraphics[scale=0.34]{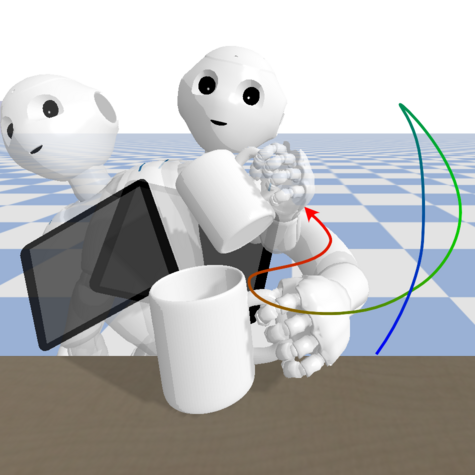}}
        \caption{Pepper}
        \label{pepper-mug}
    \end{subfigure}
    \caption{Examples of grasping behaviors found for each robot.}
    \label{traj_rob}
\end{figure}

A possible solution is to decompose the task and start by estimating object pose before determining where to grasp it \cite{kleeberger2020survey}, but these approaches require to know object models. 
Another approach consists in exploiting available knowledge about the end-effector. The features of a parallel-jaw gripper, for instance, can be used to limit the number of grasping candidates \cite{kraft2010development,mahler2017dex,depierre2018jacquard}. Unfortunately, this strategy does not easily transfer to other kinds of end-effectors. Another strategy consists in constraining the grasping movements to an easier top-down grasping motion \cite{levine}, but, by construction, it limits the range of possibilities. Human demonstrations can also be used to guide the learning process \cite{lenz2015deep,sharma2018multiple,zhang2018deep,song2020grasping}, but generating large enough datasets is costly and would need to be repeated for every new robot.



In this work, we introduce a method to autonomously build grasping datasets. The method can be applied to any kind of robot arm and end-effectors with negligible adaptation. It has been tested in simulation on grippers (Kuka Iiwa and Baxter robots) as well as on a multi-fingered  hand (Pepper robot). The transfer to a real robot has been studied on the Baxter. The proposed approach relies on a Quality Diversity algorithm \cite{pugh2016quality,cully2017quality} built on top of Novelty Search \cite{ns}. Previous works with these algorithms have generated repertoires of legged-robots movements \cite{nature} or ball throwing and joystick manipulation movements on a Baxter \cite{kim}. Novelty Search has been shown to have an efficient exploration ability as it samples uniformly in a given behavior space \cite{doncieux2019novelty,inevitable}. To deal with the sparsity of successful grasping movements, we introduce an improvement of its exploration skill through the simultaneous management of multiple behavior spaces. It drastically improves the number of grasping movements discovered. The approach generates sets of open-loop grasping movements for a given object at a given location (Fig.~\ref{traj_rob}). It cannot be used to directly grasp unknown objects at any location, but can generate large-scale grasping datasets for deep learning approaches. Although the grasping movements are generated and tested in simulation, we show that the size and diversity of the dataset result in a large number of policies that transfer to the real world.

\section{RELATED WORK}



Large scale data collection can rely on human demonstrators observation \cite{human_d, human_d2}, but these methods are hard to scale since they require large human resources, are biased by human semantic priors and are specific to one type of robotic arm.  Automating the data collection is challenging given the sparsity of grasping movements, but it has been achieved in restricted planar grasp conditions \cite{self_sup, levine}. These restrictions are sufficient to get a successful grasp rate of 10\% to 30\%, which allows to collect a sufficient dataset to train a neural network achieving grasp control from real video images \cite{levine}. Other approaches focus on parallel-plate grippers to define an autonomous grasping movement generator \cite{depierre2018jacquard}. The approach introduced here does not constrain grasping movements and can be used directly on any kind of robot.


Discovering diverse solutions is an important challenge in a variety of Machine Learning fields. Novelty-based evolutionary methods \cite{ns,pugh2016quality,map}, as well as related goal exploration process methods \cite{imgep}, were designed to tackle this problem and illuminate search spaces.
Novelty search, one of these approaches \cite{ns}, was actually shown to tend towards a uniform sampling in a user-defined behavior space \cite{doncieux2019novelty, inevitable}, a property that is hard to get given the complexity of the mapping the policy parameter space and the behavior space. These methods have been used to generate repertoires of primitive actions for robot locomotion \cite{map,nature} or simple manipulations with a robot arm \cite{kim}. The generated repertoires have already been used to bootstrap deep learning approaches \cite{gan,keller_model-based_2020}. We propose to extend novelty-based repertoire generation methods with exploration of multiple behavior spaces such as in \cite{multiqd1,multiqd2} but with an additional focus on setups with very sparse interactions. We then apply it to grasping movements generations.

\section{METHOD}

Fitness shaping is a convenient mean to turn a sparse reward into a dense reward in order to facilitate exploration \cite{ng1999policy}. Novelty-based algorithms algorithms are not driven by reward and therefore do not suffer from the sparse reward problem, but if the behaviors they look for are very sparse in the space of all possible behaviors, as is the case for grasping, a similar issue arises as the search will have trouble bootstrapping and discovering relevant solutions. In the previously studied robotics tasks such as locomotion, navigation and ball throwing, all policies (even randomly generated ones) yield a valid behavior: the robot reaches some final position, or the ball reaches some location. The first discovered behaviors are usually very simple (such as remaining very close to the start position or just dropping the ball), but they allow the algorithm to bootstrap and then iteratively discover more novel and higher quality policies. By contrast, trying to bootstrap grasping policies with random generation in a large policy space with little or no prior knowledge will mostly produce behaviors that do not engage the target object at all.

To mitigate this issue, we introduce an approach for \textit{behavior shaping} that has a goal similar to fitness shaping, but in behavior spaces. The idea is to introduce new behavior spaces whose exploration aims at facilitating the generation of ``target'' behaviors, in a curriculum-like way. Building upon the ability of Novelty Search to continuously generate novel solutions \cite{inevitable}, Novelty Search with Multiple Behavior Spaces (NSMBS) extends it to take into account multiple behavior spaces and thus propose a framework in which behavior shaping can be implemented.


\subsection{Definitions and notations}

In this paper, we consider policies $\pi_\theta$ parameterized by a vector $\theta \in \Theta \subset \mathbb{R}^{n_g}$, with $n_g$ the policy dimension. The policies, as well as the transition model of the robot and environment, are assumed deterministic. Novelty-based methods rely on the policies' \textit{behavior characterization}, which results from mapping the trajectory of the robot when applying the policy from an initial state $s_0$ into a smaller dimension vector in a carefully selected \textit{behavior space} \cite{doncieux2019novelty}. In the following and as usual in novelty-based algorithms, we assume that the initial state is fixed, and the behavior therefore only depends on the policy parameter $\theta$.

NSMBS proposes a framework to explore in multiple behavior spaces, here called behavior components. A policy is thus be described by $n_b$ behavior components. While in classic novelty-based algorithms, each policy is always associated to a behavior characterization, NSMBS relaxes this assumption and some policies may have undefined behavior components. It allows us to characterize grasping behaviors, for instance, and give an undefined grasping behavior component to policies that do not succeed in grasping the object. To this end, an eligibility criterion is associated to each behavior component. This gives rise to the following notations:
\begin{itemize}
    \item $\mathcal{B}_i \subset \mathbb{R}^{n_{B_i}}$ is the $i$-th behavior component space;
    \item $\xi_i(\theta)$ is the eligibility criterion associated to $\mathcal{B}_i$. If $\xi_i(\theta)=\text{True}$, the corresponding behavior is defined.
    \item $b_i(\theta) \in \mathcal{B}_i  \cup \{\varnothing\}$ is the $i$-th component of the robot behavior when it follows  policy $\pi_\theta$. $b_i(\theta) = \varnothing$ i.f.f. $\xi_i(\theta)=\text{False}$;
    \item $n_b$ is the number of behavior components.
\end{itemize}

Although some recent work try to automatically learn the behavior spaces \cite{paolo2020unsupervised,cully2019autonomous}, here the behavior components are supposed to be known and given by the experimenter. 

\subsection{NSMBS}

NSMBS (Algorithm~\ref{alg:NSMBS}) derives from Novelty Search \cite{ns,doncieux2019novelty}. As for other evolutionary algorithms, an initial population of solutions is randomly generated, evaluated, and a parent set is selected. The parents are then copied and modified by mutation and crossover operators to get new solutions, that are then evaluated and selected to constitute the next generation of parents. In Novelty Search, an archive of explored behaviors is also maintained, and the selection process relies on maximizing a novelty objective, which is the average distance to the $K$-nearest neighbors in the behavior space among the archive and current population.

In NSMBS, the outcome of a policy $\pi_\theta$ is described by a list of behavior components: $\left(b_1(\theta), b_2(\theta), \ldots, b_{n_b}(\theta)\right)$. The selection process in charge of finding the individuals of the next generation relies on these elements (Algorithm \ref{alg:multibcsel}). Policies are selected one by one from a set of individuals $\mathcal{S}$ initialized with the current population and offspring. The individual selection starts by randomly choosing a behavior component among the ones that are defined at least for one policy in $\mathcal{S}$. The most novel individual in the selected component is then selected; it is removed from $\mathcal{S}$ and the process starts again until the new population is filled.

\begin{algorithm}
\small
\SetAlgoLined
\SetKwInOut{Input}{Input}
\Input{population size $\mu$, number of generations $G$, number of offsprings $\lambda$, evaluation function $eval()$, number of neighbours for novelty computation $k$}
\KwResult{archive of individuals}
 $\text{pop} \leftarrow generateRandomPopulation(\mu)$ \;
 $\text{archive} \leftarrow \varnothing$ \;
 \For{$a$ in $\text{population}$}{
  $a.bd\leftarrow eval(a)$ \;
 }
 $\text{gen} = 0$ \;
 \While{gen $< G$}{
  $\text{parents} \leftarrow selectParents(pop, \lambda)$ \;
  $\text{offspring} \leftarrow operate(\text{parents})$ \;
  \For{$a$ in $\text{offspring}$}{
   $a.bd \leftarrow eval(a)$ \;
  }
  $\text{refSet} \leftarrow \text{pop} \cup \text{offspring} \cup \text{archive}$ \;
  \For{$a$ in $\text{pop} \cup \text{offspring}$}{
   $ a.nov \leftarrow getNov(a, \text{refSet}, k)$ \;
  }
  $\text{archive}.add(randomSample(\text{offspring}))$ \;
  $\text{pop} \leftarrow multiBCSel(\text{pop} \cup \text{offspring}, \mu)$ \;
  $ \text{gen} = \text{gen} + 1$ \; 
 }
 \caption{NSMBS}
 \label{alg:NSMBS}
\end{algorithm}

\begin{algorithm}
\small
\SetAlgoLined
\SetKwInOut{Input}{Input}
\Input{set of individuals $\mathcal{S}$, number of individuals to select $\mu$}
\KwResult{set of individuals $\mathcal{P}$}
 $\mathcal{P} \leftarrow \varnothing$ \;
 \While{$size(\mathcal{P})<\mu$}{
    $i \leftarrow randomSelComponent(\mathcal{S})$ \;
    $\mathcal{T} \leftarrow \{\theta \in \mathcal{S} \mid \xi_i(\theta)=\text{True} \}$ \;
    $\text{chosen} \leftarrow \argmax\limits_{\theta \in \mathcal{T}} {nov_i(\theta)}$ \;
    $\mathcal{P} \leftarrow \mathcal{P} \cup \{\text{chosen}\}$ ; $\mathcal{S} \leftarrow \mathcal{S} - \{\text{chosen}\}$ \;
 }
 \caption{multiBCSel}
 \label{alg:multibcsel}
\end{algorithm}

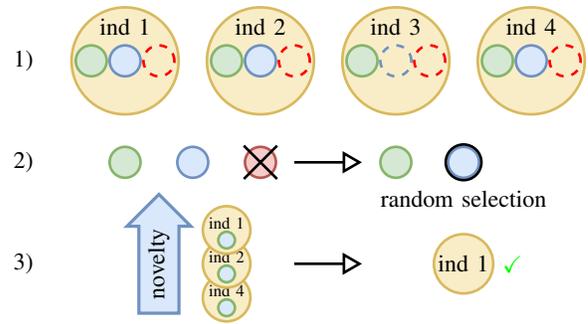
\begin{figure}[thpb]
    \centering
    \definecolor{fillind}{RGB}{249,237,200}
    \definecolor{drawind}{RGB}{216, 185, 93}
    \definecolor{fillgreen}{RGB}{213, 232, 212}
    \definecolor{drawgreen}{RGB}{134, 181, 105}
    \definecolor{fillblue}{RGB}{218, 232, 252}
    \definecolor{fillbluemid}{RGB}{126, 178, 252}
    \definecolor{fillbluedeep}{RGB}{0, 105, 252}
    \definecolor{drawblue}{RGB}{112, 145, 193}
    \definecolor{fillred}{RGB}{246, 204, 202}
    \definecolor{drawred}{RGB}{184, 86, 82}
    \newcommand{\ind}[3]{
    \node [circle, draw=drawind, fill=fillind, line width=1, minimum size=45, path picture={
        \draw
              (-0.5, 0) node [circle, draw=drawgreen, minimum size=13, fill=fillgreen] {}
              (0, 0) node [circle, draw=drawblue, minimum size=13, #1] {}
              (0.5, 0) node [circle, draw=red, minimum size=13, dashed] {}
              (0, 0.5) node [] {#2};
        }] (#2) at (#3,0) {};
    }
    \newcommand{\smallind}[2]{
        (-1.5,#2) node [circle, draw=drawind, fill=fillind, line width=1, minimum size=20, path picture={
            \draw
              (0, -0.15) node [circle, draw=drawgreen, minimum size=7, inner sep=0, fill=fillblue] {}
              (0, 0.1) node [] {\scriptsize #1};
        }] {}
    }
    \begin{tikzpicture}[scale=0.9, every node/.style={transform shape}]
        \tikzmath{\vspace=1.5;}
        \ind{fill=fillblue}{ind 1}{-3}
        \ind{fill=fillblue}{ind 2}{-1}
        \ind{dashed}{ind 3}{1}
        \ind{fill=fillblue}{ind 4}{3}
    
    \draw [line width=1, minimum size=13]
        (-3,-\vspace) node [circle, draw=drawgreen, fill=fillgreen] {}
        (-2,-\vspace) node [circle, draw=drawblue, fill=fillblue] {}
        (-1,-\vspace) node [circle, draw=drawred, fill=fillred] {}
        (-1,-\vspace) node [cross out, draw] {}
        
        (1,-\vspace) node [circle, draw=drawgreen, fill=fillgreen] {}
        (2,-\vspace) node [circle, draw=drawblue, fill=fillblue] {}
        (2,-\vspace) node [circle, draw, minimum size=15, label={below:random selection}] {}
        (-0.5,-\vspace) edge[->, >=open triangle 60, line width=1] (0.5,-\vspace);

    \draw []
        (-2.5,{-2*\vspace}) node[single arrow, draw=drawblue, very thick, fill=fillblue, minimum width = 10pt, single arrow head extend=3pt, minimum height=50, single arrow head extend=5pt, rotate=90] {novelty}
        \smallind{ind 4}{{-2*\vspace-0.5}}
        \smallind{ind 2}{{-2*\vspace}}
        \smallind{ind 1}{{-2*\vspace+0.5}}
        (-0.5,{-2*\vspace}) edge[->, >=open triangle 60, line width=1] (0.5,{-2*\vspace})
        (2, {-2*\vspace}) node [circle, draw=drawind, fill=fillind, line width=1, minimum size=1, inner sep=1, label={right:\color{green}\checkmark}] {ind 1};
    
    \draw []
        (-4.5, 0) node {1)}
        (-4.5, -\vspace) node {2)}
        (-4.5, {-2*\vspace}) node {3)};
  \end{tikzpicture}
  \caption{Description of the \textit{multiBCSel()} method for 4 individuals and 3 behavior components. The third (red) component cannot be selected because there is no eligible individual. Among the two remaining components, the second (blue) one is randomly selected, and the eligible individuals (inds 1, 2 and 4) are sorted according to $nov_2$. The most novel individual, ind 1, is selected.}
  \label{sel}
    \vspace{-5pt}
\end{figure}

\begin{itemize}
    \item \textit{generateRandomPopulation} generates a population of size $\mu$: $\{\theta_0, ... \theta_{\mu-1}\}$ by uniform sampling;
    \item \textit{eval} evaluates an individual's policy $\theta$ and outputs its behavior components $bd=\left(b_1(\theta), b_2(\theta), \ldots, b_{n_b}(\theta)\right)$, with $b_i(\theta) = \varnothing$ i.f.f. $\xi_i(\theta)=\text{False}$
    \item \textit{selectParents} is a random selection process of $\lambda$ individuals in $\text{pop}$;
    \item \textit{operate} copies the parents and applies gaussian bounded mutation and crossover operators to generate offsprings;
    \item \textit{getNov} computes a list of per component novelties $\text{nov} = \left(\text{nov}_1, \text{nov}_2, \ldots, \text{nov}_{n_b}(\theta)\right)$ for an individual. $\text{nov}_i = \sum\limits_{k = 0}^{K-1}\left(dist\left(b_i(\theta), b_i(\theta_k)\right)\right)$ where $b_i(\theta_k)$ are the behavior components of the $K$ nearest neighbors of $b_i$ in $\text{refSet}$ ($\varnothing$ values are ignored). 
    If $b_i(\theta) = \varnothing$, $\text{nov}_i$ is undefined. 
    \item \textit{multiBCSel} is the main selection process, described in Fig.~\ref{sel} and Algorithm~\ref{alg:multibcsel}.
   \item \textit{randomSelComponent} selects a behavior component by uniform selection among components for which at least one individual is eligible.
\end{itemize}


\section{Experimental Setup}

\subsection{Simulated environments description}

In order to demonstrate the algorithm's ability to generate rich, diverse grasping movements for multiple robots, we rely on simulated environments 
using the Gym framework \cite{gym}. A robot is placed in front of a table, where an object is spawned at a fixed position. Three robots are used: A Rethink Robotics Baxter using only the left arm, with 7 degrees of freedom and a parallel gripper; A Kuka Iiwa arm with 7 degrees of freedom and a two-fingers clamp gripper, both simulated using pyBullet~\cite{coumans2019}; and a SoftBank Robotics Pepper, using only the left arm, with 5 degrees of freedom and a multi-fingered hand, simulated using Qibullet~\cite{qibullet}. The episode's length $T$ is adapted to each robot and simulator's dynamics, all other experimental hyperparameters being the same between the three robots. Five target objects are used: a simple \SI{5}{\centi\meter} cube, a simple \SI{5}{\centi\meter} ball, a miniature plastic bowling pin, a mug with a handle, and a gamepad. 

The policies are defined by three waypoints in the joint space, giving the robot pose at $T/3$, $2T/3$ and $T$. Smooth motions are interpolated using third order polynomials. An extra parameter $t_{grasp}$ defines the time when the gripper closes. The policy dimension $n_g$ is therefore $3 \times n_{dof} + 1$.

Besides the length of the episode $T$, we define $t_{touch}$ the time when the gripper first touches the object (undefined if the object is not touched), $obj^{pos}_t$ and $grip^{pos}_t$ the position of resp. the object and the gripper at time $t$ in the Cartesian space ($\mathbb{R}^3$), and $grip^{or}_t \in \mathbb{R}^4$ the orientation of the gripper at $t$, defined as a quaternion.

\subsection{Behavior Descriptors}

We consider the following four behavior descriptors:
$b_1(\theta) = obj^{pos}_T$, 
$b_2(\theta) = grip^{or}_{T/2}$,
$b_3(\theta) = grip^{pos}_{t_{touch}}$ and
$b_4(\theta) = grip^{or}_{t_{touch}}$.
Each $b_i$ pushes the exploration in one direction and generates a useful incentive: $b_1$ pushes for trajectories that move the object around, $b_2$ promotes diverse movements during exploration, $b_3$ and $b_4$ explicitly push towards diverse grasping poses.

$b_1$ is always eligible. $b_2$ is only eligible if the gripper touched the object (we only search diverse trajectories that engage the object). $b_3$ and $b_4$ are only eligible if the object was successfully grasped. A successful grasp is defined as a trajectory where the gripper touches the object shortly after it is closed while the object is on the table, and that ends with the object in a stable position in the air, with minor conditions on penetrations to avoid unrealistic grasps.  

\subsection{Measures}
\label{ssec:measures}

To monitor the progress of the exploration and the resulting diversity, we use the following metrics:
\begin{itemize}
\item The number of successful grasping solutions found, and their \textit{coverage} of the behavior space $\mathcal{B}_4$ \cite{inevitable}. This gives an estimation of the raw success of the repertoire generation process.
\item The \textit{sample efficiency}, defined as the proportion of all the evaluations that resulted in a successful grasp. This measures the ability to generate a rich repertoire while minimizing the computational cost.
\end{itemize}
In order to compare our method to the state of the art, we introduce two other measurements:
\begin{itemize}
\item The \textit{first success generation}, which is the number of iterations after which the algorithm first discovers a successful grasp. This evaluates the ability of an algorithm to quickly bootstrap in this sparse, difficult task.
\item The \textit{successful run rate}, which measures the proportion of runs that generated at least a single grasping movement. This measures the reliability of an algorithm.
\end{itemize}






\section{RESULTS}

\subsection{Grasping synthesis on multiple robots}
\label{ssec:multirobot}

The NSMBS algorithm is first run on all the three robots, with the mug object, for 1000 generations, a population size $\mu = 100$ individuals, an offspring size $\lambda = 50$ individuals, and a value of $K=15$ for the KNN algorithm. The search is repeated 10 times for each condition.

The generated repertoires are very different between robots, with sizes of $13148 \pm 4574$ successful grasps for the Baxter, $3490 \pm 4706$ for the Pepper, and $3929 \pm 2262$ for the Kuka. This reflects the variable difficulty of the grasping task with different robots, the high variance for the Pepper and Kuka robots being due to the presence of failed runs where no grasping was discovered. Despite this, the search finds a variety of grasping behaviors for all robots, some of which are shown in Fig.~\ref{traj_rob}.

\begin{figure}[ht]
    \centering
    
    \begin{subfigure}[b]{0.47\columnwidth}
        \centerline{\includegraphics[scale=0.4]{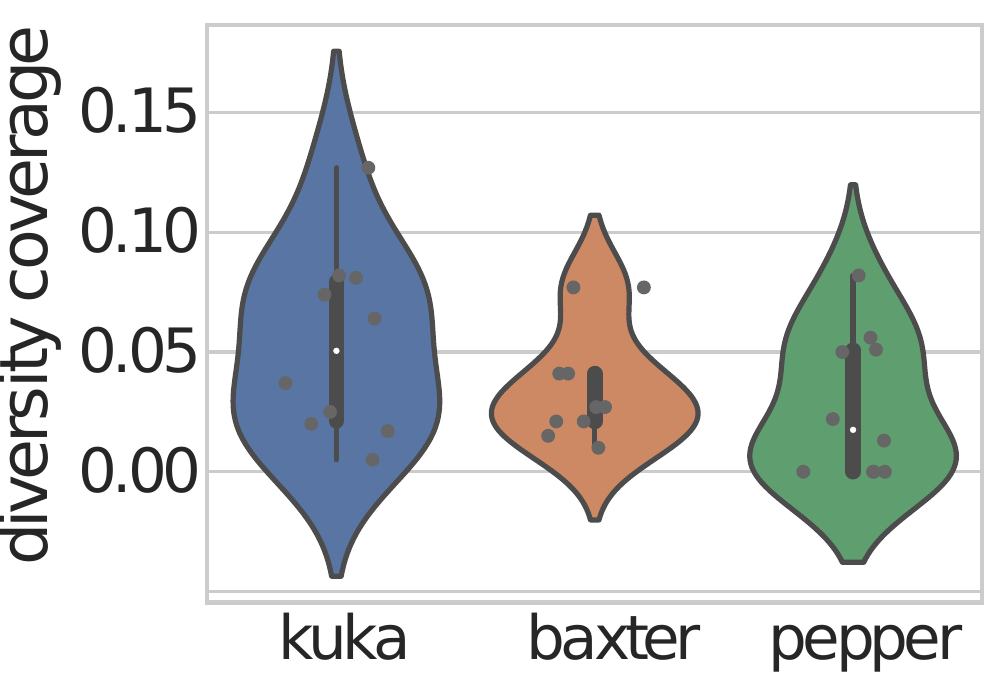}}
        \caption{Diversity coverage of NSMBS for all robots.}
        \label{robots-mug-overall}
    \end{subfigure}
    \begin{subfigure}[b]{0.47\columnwidth}
        \centerline{\includegraphics[scale=0.38]{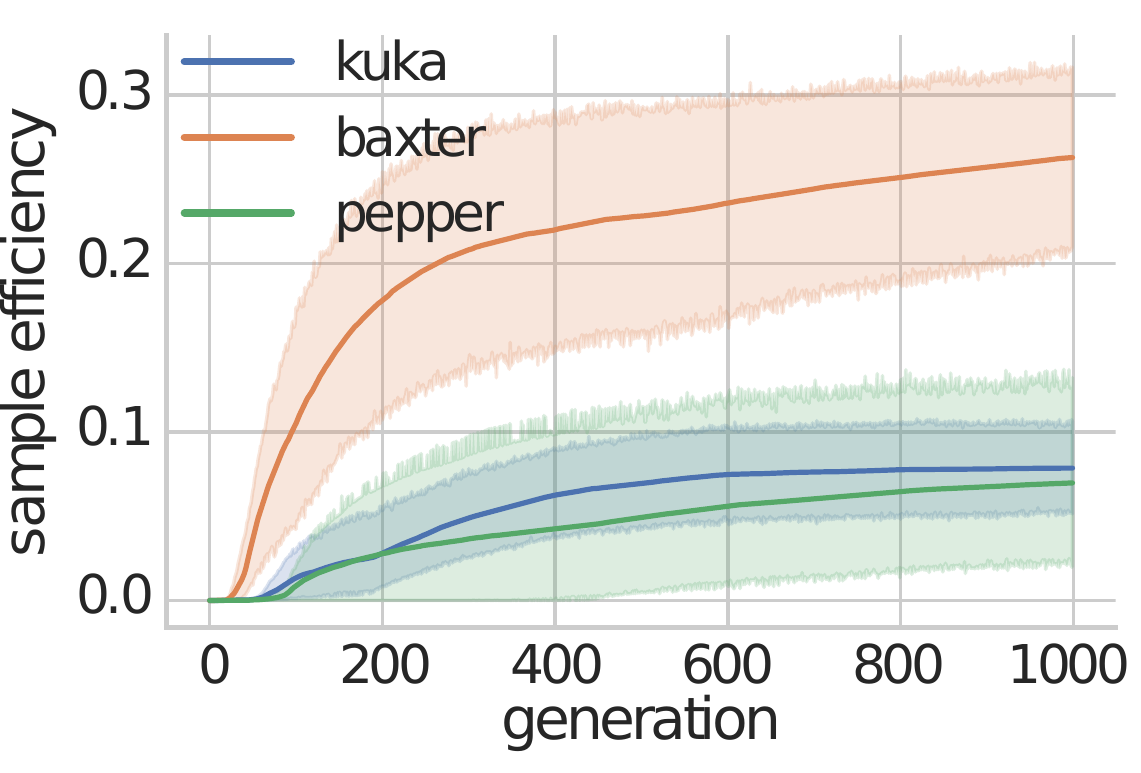}}
        \caption{Evolution of sample efficiency during the search.}
        \label{robots-mug-per-generation}
    \end{subfigure}
    
    \caption{Coverage and sample efficiency of the search for grasping behavior with NSMBS for the three robots and the mug object.}
    \label{robots-mug}
\end{figure}

Fig.~\ref{robots-mug} shows the final diversity coverage and the evolution of the sample efficiency for the three robots. We can see that the algorithm quickly manages to bootstrap and discover grasping motions, and then leverages those existing motions to improve diversity, as shown by the constantly growing sample efficiency during the process.






Furthermore, we can sort the ways the mug object can be grasped in four different categories, which allows us to define four grasping styles, not all of them available to all robots due to their design:
\begin{itemize}
    \item Grasping the handle (\texttt{handle} style): this grasping style is possible on all three robots;
    \item Grasping the mug from the top, with part of the gripper inside the mug and the other outside (\texttt{in-out} style): only the Kuka and Baxter robots are capable of this grasping style, the Pepper robot's weaker hand being unable to hold the mug this way;
    \item Grasping the mug from the sides, with the gripper grasping the outside of the mug (\texttt{out} style): only the Kuka robot is capable of this grasping style, thanks to its much larger gripper;
    \item Grasping the mug from the inside, by inserting the entire gripper (\texttt{in} style): only the Baxter robot with its thin parallel gripper is capable of this grasping style.
\end{itemize}

\begin{figure}[thpb]
    \centering
    \includegraphics[scale=0.3]{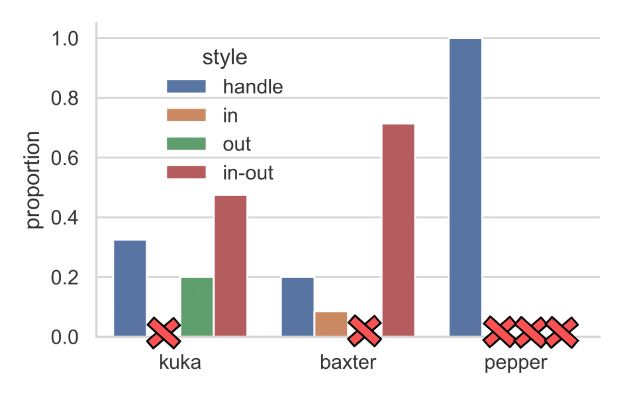}
    \caption{Mug grasping styles proportions (10 runs, 1000 generations). Red cross means grasp style not generated for this robot.}
    \label{mug-grasping-style}
    \vspace{-10pt}
\end{figure}

We categorize the discovered grasping motions on all three robots. The results (Fig.~\ref{mug-grasping-style}) show that NSMBS is general enough to fully make use of the capabilities of each robot, discovering all the possible styles for all robots.

\subsection{Comparative analysis}
\label{ssec:comparison}
In order to highlight the benefits of NSMBS, we compare the performance with the following variants and baselines:
\begin{itemize}
    \item Our full NSMBS algorithm (NSMBS);
    \item The same algorithm without the object touching BD (NSMBS no BD2) and without the grasping position BD ((NSMBS no BD3);
    \item A classic Novelty Search (NS) algorithm~\cite{ns} using a single behavior descriptor, which is built by concatenating the four $b_i$ from NSMBS. If one of the components is not eligible, its values are set to 0.
    \item A Map-Elites algorithm~\cite{map}, using the same concatenated behavior descriptor as NS with 1000 CVT cells \cite{cvt}. Map-Elites uses a quality measure; it is not important here and set to a simple energetic criterion;
    \item A random search baseline where a similar number of individuals are generated by random uniform sampling in policy parameter space.
\end{itemize}

\begin{figure}[ht]
    \centering
    \hfill
    \begin{subfigure}[b]{0.47\columnwidth}
        \centerline{\includegraphics[scale=0.62]{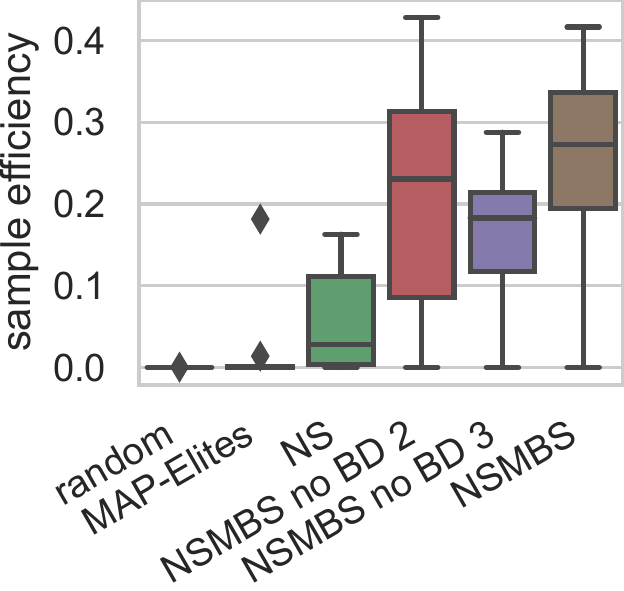}}
        \caption{Overall sample efficiency of each algorithm.}
        \label{sample-efficiency-algorithms}
    \end{subfigure} \hfill
    \begin{subfigure}[b]{0.47\columnwidth}
        \centerline{\includegraphics[scale=0.62]{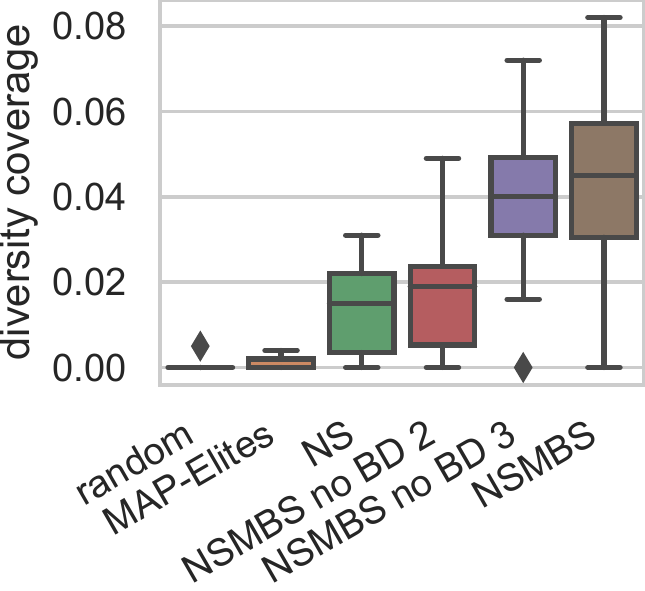}}
        \caption{Final diversity coverage of each algorithm.}
        \label{diversity-algorithms}
    \end{subfigure}
    \hfill
    %
    
    \caption{Sample efficiency and diversity coverage of the search for grasping behavior on the Baxter robot after 2000 generations.}
    \label{algos-comparison-general}
    \vspace{-10pt}
\end{figure}

The algorithms are used on the Baxter environment with the mug object, with the same hyperparameters as above except the process is run for 2000 generations. Each condition is repeated 20 times. The main results are shown in Fig.~\ref{algos-comparison-general}. The NSMBS variants overperforms the baselines, yielding higher sample efficiency and diversity coverage.
NS discovers some grasping motions, but is generally much less sample-efficient, resulting in a lower coverage. Map-elites is generally unsuccessful ; this may be due to the absence of a population combined with a random selection and a very large composite behavior space, which fails to focus the evolutionary budget on promising individuals~\cite{coninx2021youth}.

\begin{figure}[ht]
    \centering
    \begin{subfigure}[t]{0.48\columnwidth}
        \centerline{\includegraphics[scale=0.5]{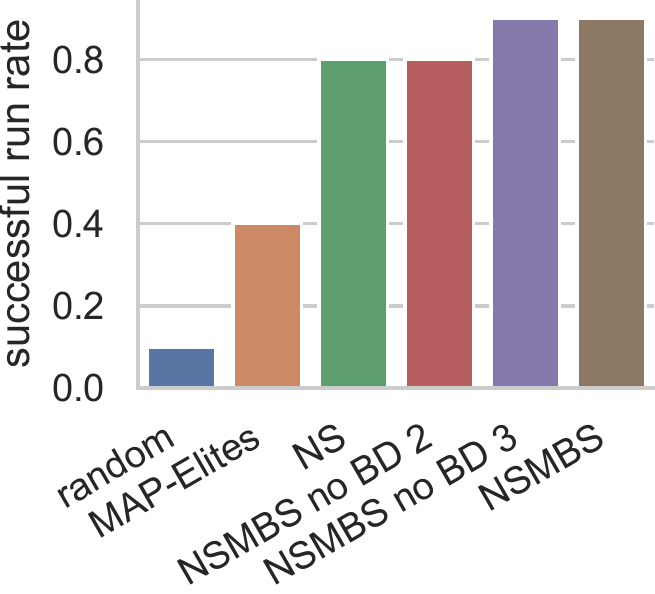}}
        \caption{Proportion of successful runs (i.e.runs that discover at least one grasp within 2000 generations) for each algorithm.}
        \label{success-algorithms}
    \end{subfigure} \hfill
    \begin{subfigure}[t]{0.48\columnwidth}
        \centerline{\includegraphics[scale=0.6]{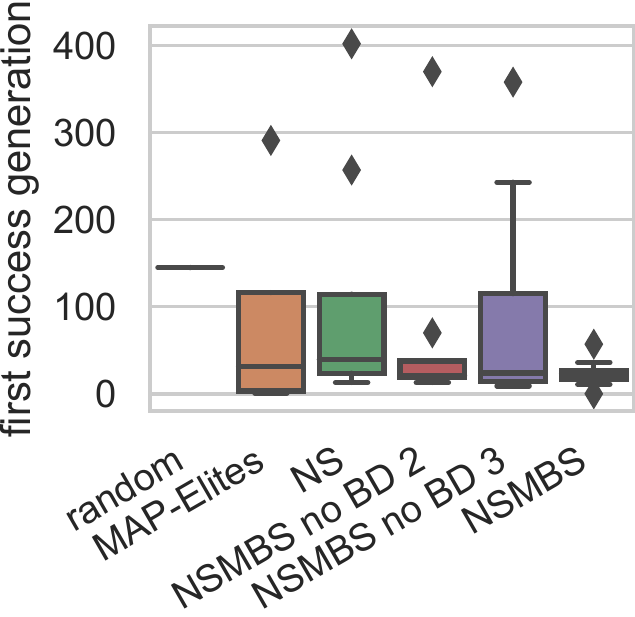}}
        \caption{Generation at which the first grasp is found, for each algorithm. Unsuccessful runs are ignored.}
        \label{first-grasp-algorithms}
    \end{subfigure}
    
    
    
    \begin{subfigure}[b]{0.4\textwidth}
        \centerline{\includegraphics[scale=0.6]{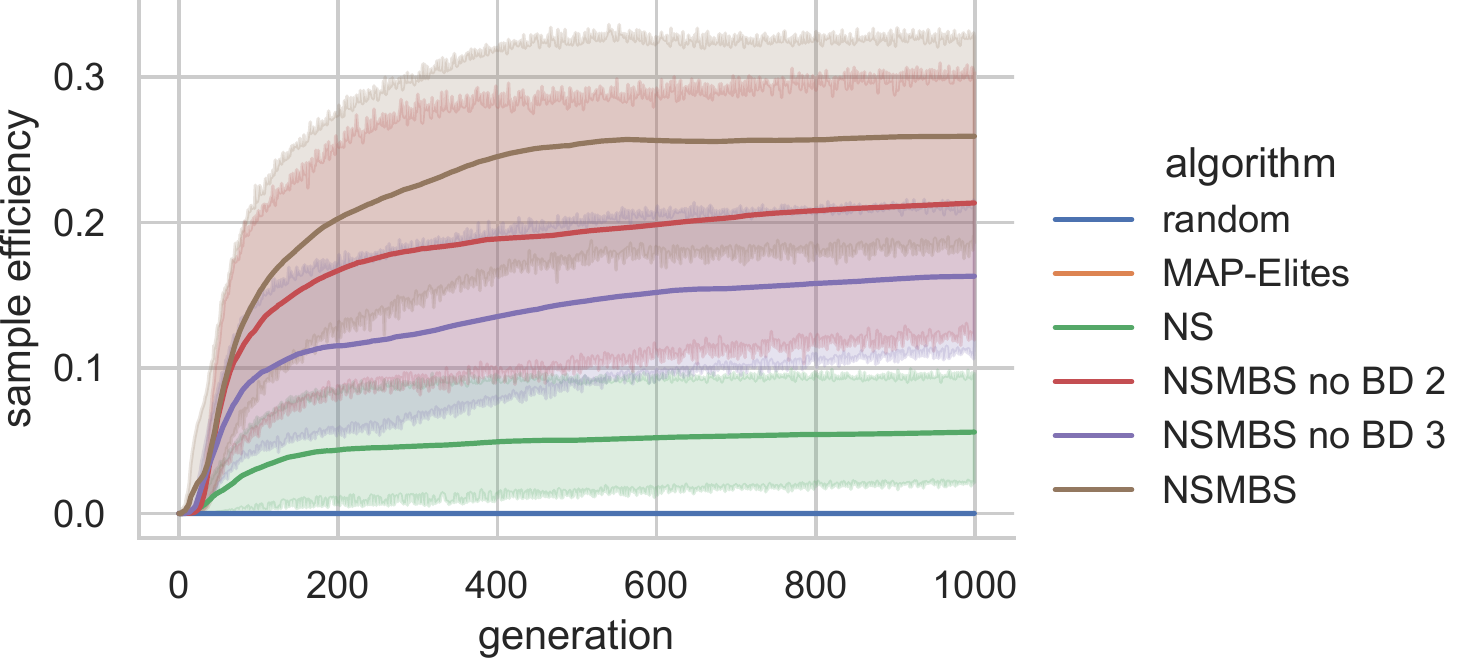}}
        \caption{Evolution of sample efficiency during the runs.}
        \label{algo-sample-evolution}
    \end{subfigure}

    \caption{Proportion of successful runs, bootstrap time for successful runs, and evolution of the  sample efficiency for each algorithm (map-elites hidden behind random).}
    \label{algos-comparison-bootstrap}
\end{figure}

Fig.~\ref{algos-comparison-bootstrap} gives further insights into the benefits of our behavior shaping technique. Although most NS and NSMBS runs are eventually successful at generating at least some grasping behaviors (Fig.~\ref{success-algorithms}), the full NSMBS algorithm with all the behavior components does so earlier in the evolutionary process (Fig.~\ref{first-grasp-algorithms}). It also quickly achieves and maintains higher sample efficiency (Fig.~\ref{algo-sample-evolution}), with 25\% of evaluated individuals (in average) being successful grasps after only 400 generations.

\subsection{Transfer on real-world robot}
\label{ssec:real}
\begin{figure*}[h!]
    \centering
    \begin{subfigure}[b]{0.18\textwidth}
        \centerline{\includegraphics[scale=0.36]{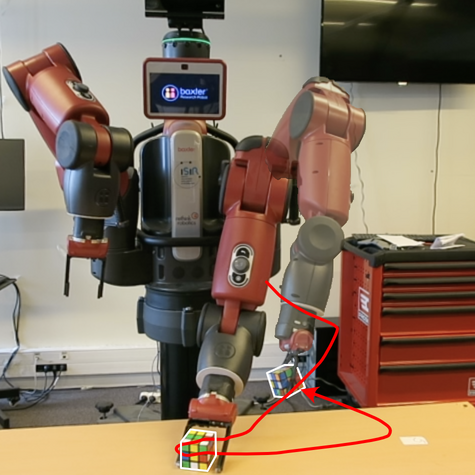}}
        \caption{Cube}
        \label{}
    \end{subfigure}
    \begin{subfigure}[b]{0.18\textwidth}
        \centerline{\includegraphics[scale=0.36]{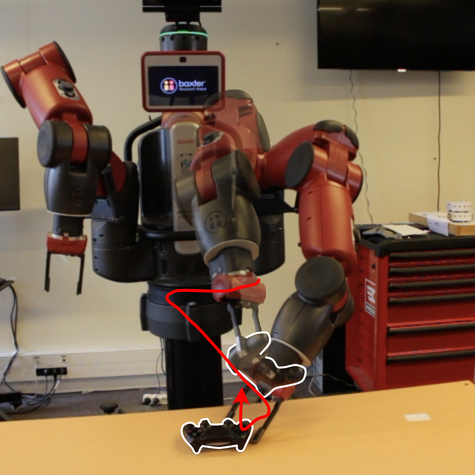}}
        \caption{Gamepad}
        \label{}
    \end{subfigure}
    \begin{subfigure}[b]{0.18\textwidth}
        \centerline{\includegraphics[scale=0.36]{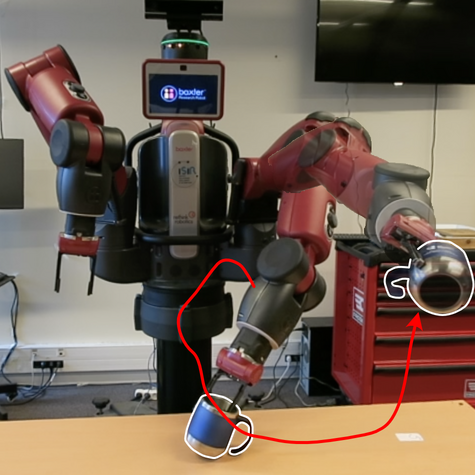}}
        \caption{Mug}
        \label{}
    \end{subfigure}
    \begin{subfigure}[b]{0.18\textwidth}
        \centerline{\includegraphics[scale=0.36]{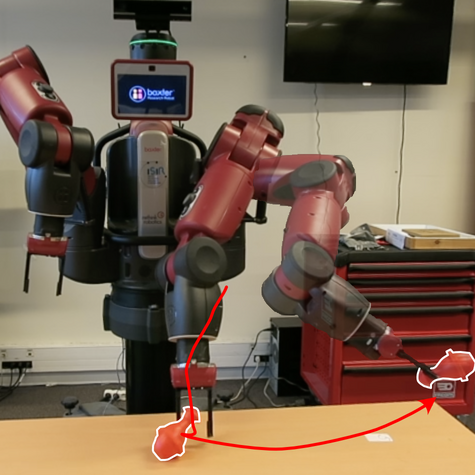}}
        \caption{Pin}
        \label{}
    \end{subfigure}
    \begin{subfigure}[b]{0.18\textwidth}
        \centerline{\includegraphics[scale=0.36]{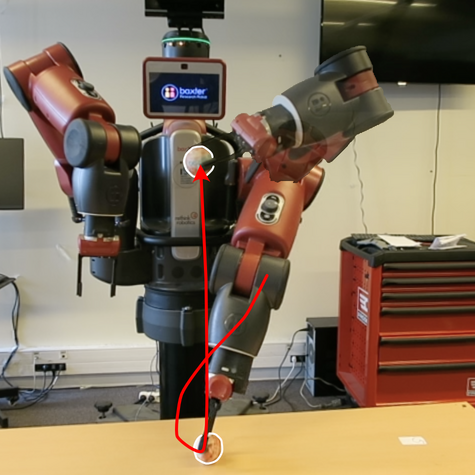}}
        \caption{Sphere}
        \label{}
    \end{subfigure}
    
    \caption{Grasping example on real Baxter robot for each object. The end effector trajectory, the pose at which the robot grasps the object and the final robot and object pose are shown.}
    \label{baxter-real-trajectories}
    \vspace{-10pt}
\end{figure*}

The grasping motion repertoires are only relevant if at least some of the generated policies transfer to the real world. We therefore generate repertoires for five objects with the Baxter robot and evaluate them in real world experiments.

\begin{figure}
    \centerline{\includegraphics[scale=0.6]{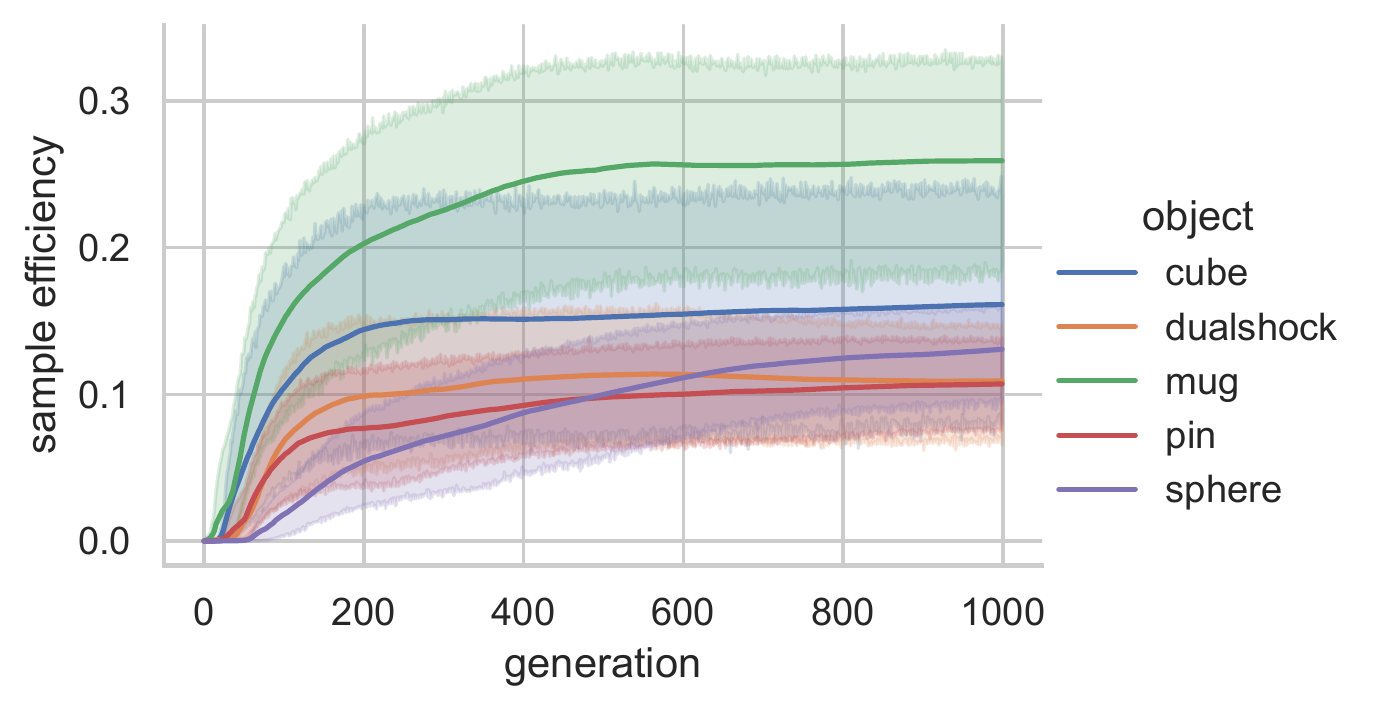}}
    \caption{Sample efficiency at each generation in simulation.}
    \label{object-sample-efficiency-per-generation}
\end{figure}

NSMBS is first run 10 times for 1000 generations in the simulated Baxter environment for each of the five objects. Repertoires are successfully generated for all objects (Fig.~\ref{object-sample-efficiency-per-generation}). Results for the mug were previously discussed in section~\ref{ssec:comparison}; sample efficiency is lower for other objects, but NSMBS still reliably generates large repertoires of successful grasps.

\begin{figure}[thpb]
    \hfill
    \begin{subfigure}[t]{0.45\columnwidth}
        \centerline{\includegraphics[scale=0.4]{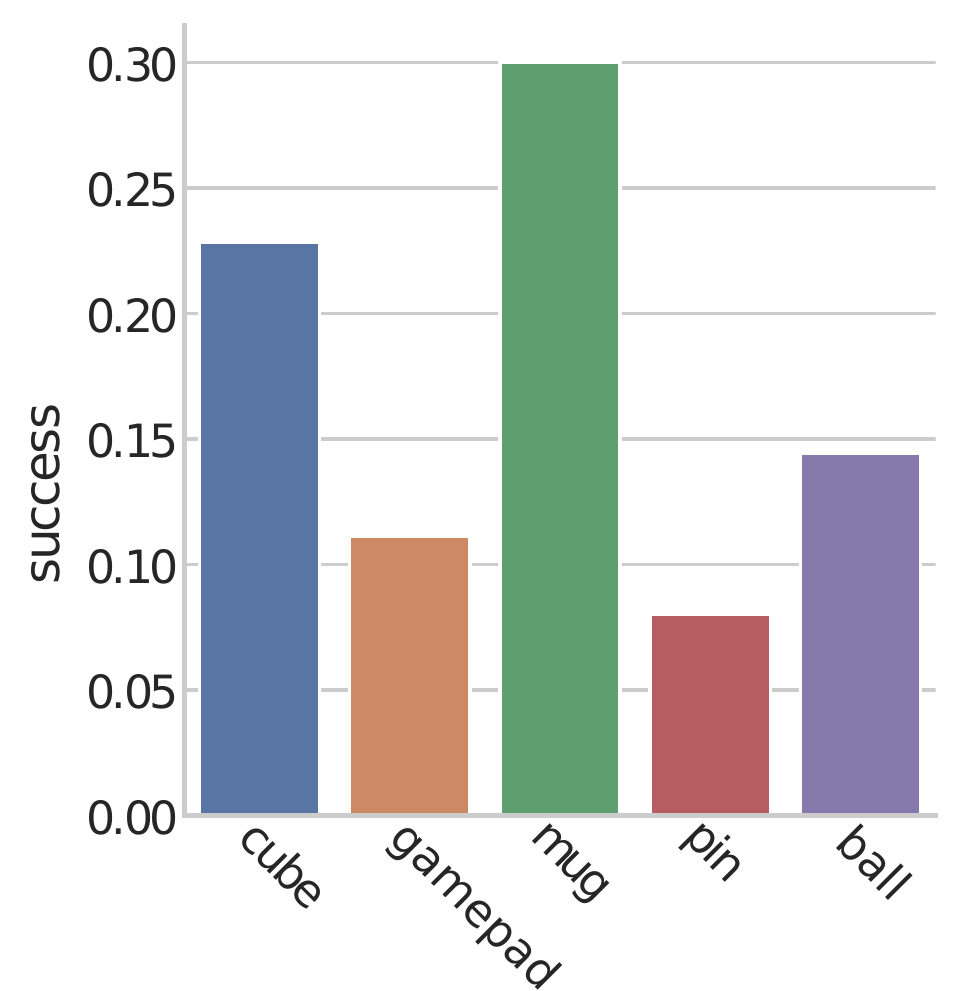}}
        \caption{Success rate of evaluated individuals in reality.}
        \label{reality-success-rate}
    \end{subfigure} \hfill
    \begin{subfigure}[t]{0.45\columnwidth}
        \centerline{\includegraphics[scale=0.4]{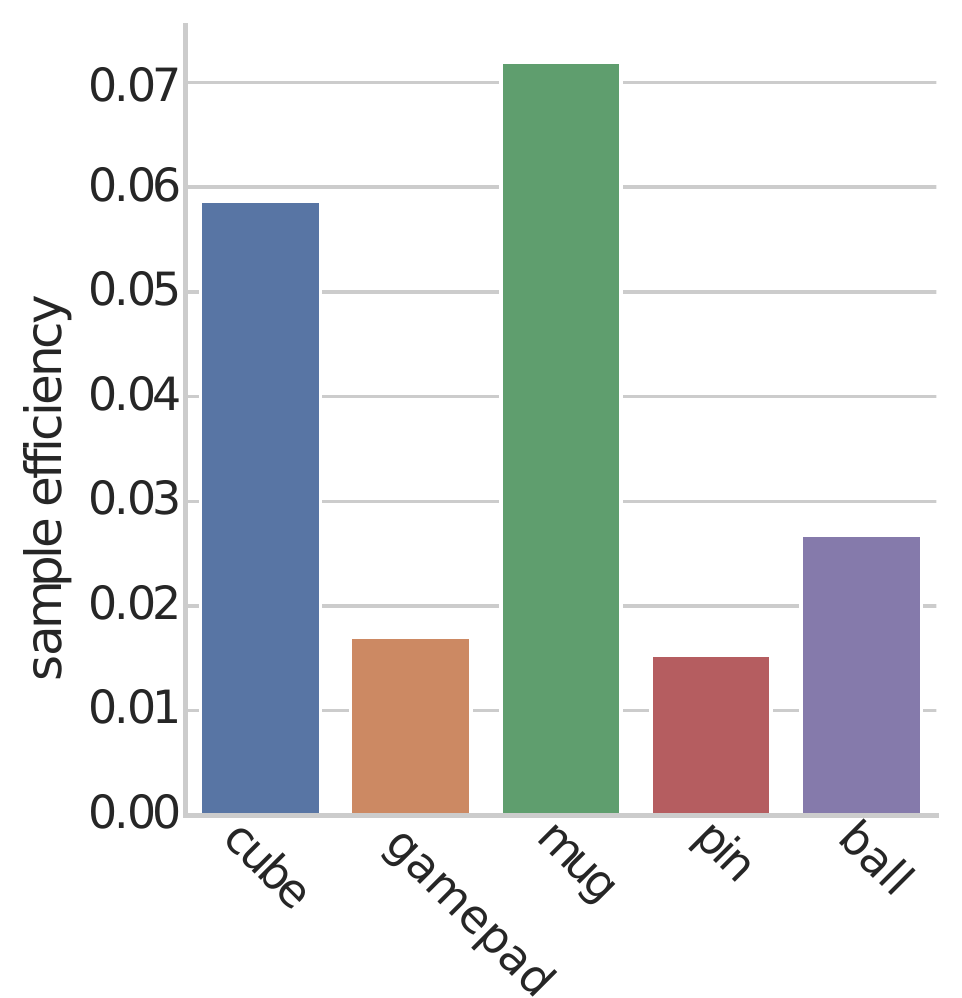}}
        \caption{Estimated sample efficiency for the full repertoire generation process in reality.}
        \label{reality-sample efficiency}
    \end{subfigure}
    \hfill
    
    \caption{Success rate and sample efficiency in reality on the robot Baxter. Repertoire generation is run 10 times per object, and 10 individuals are randomly selected from each successfully generated repertoire.}
    \label{baxter-reality}
    \vspace{-12pt}
\end{figure}

All 10 runs succeed in discovering grasping motions for the pin, 9 succeed for the mug, the sphere and the gamepad, and 7 succeed for the ball. For each successful run, 10 individuals are randomly sampled from the final repertoire, and evaluated on the robot, for a total of 70 to 100 real world evaluations depending on the object. Some successful grasps for each object are shown in Fig.~\ref{baxter-real-trajectories} and in the video in annex.

The transfer success rates are reported in Fig.~\ref{reality-success-rate}. The highest transfer rate is achieved with the mug, where \SI{30}{\percent} of policies succeeded, followed by the cube, the sphere, and finally the gamepad and the pin at \SI{8}{\percent}. It is noteworthy that this success rate is directly correlated with the sample efficiency of the process in simulation (Fig.~\ref{object-sample-efficiency-per-generation}). The lower sample efficiency for the gamepad and the pin can be explained by their characteristics : the gamepad is quite heavy (\SI{0.21}{\kilo\gram}) compared to other objects, and it has a complex shape and moving parts that are not fully represented in the simulated environment. The pin is standing, thus it can easily fall and the dynamics then differ from the simulation.

We can use these results to estimate the sample efficiency of the full repertoire generation process, i.e. the expected proportion of all the evaluations in simulation that result in a grasp that transfers to reality, by multiplying the sample efficiency in simulation (as defined in section~\ref{ssec:measures}) by the transfer success rate. Results are shown in Fig.~\ref{reality-sample efficiency}. Values range from about \SI{1.6}{\percent} for the pin to \SI{7.1}{\percent} for the mug.

\section{CONCLUSION}

The simulation results show that our NSMBS method is able to efficiently generate grasping behaviors for various robot arms, and that it outperforms the state of the art in both diversity coverage and sample efficiency. The comparison to ablated versions of the method with only some of the behavior components highlights the importance of choosing adequate behavior characterizations for the method to be able to bootstrap and learn, and the comparison to classic diversity methods showcases the ability of NSMBS to take advantage of those multiple components for behavior shaping.

Real world transfer rates range from \SI{8}{\percent} to \SI{30}{\percent} depending on the object. This is comparable to the technique used by Levine et al.~\cite{levine}, which makes strong hypotheses about the experimental setup and uses motor primitives that constrains the diversity of the possible grasps, whereas our method is applicable to a large variety of robots without any modification and promotes grasping diversity.

The real world sample efficiency ranges from \SI{1.6}{\percent} to \SI{7.1}{\percent}, which corresponds to one out of 14 to one out of 62 simulated policies being a valid grasping policy in the real world. It may seem low, but it does not prevent the generation of a large, diverse repertoire, as simulated experiments are cheap enough that more than \num{100000} evaluations can be performed in a few hours on a modern workstation, resulting in thousands of transferable grasps. An open question is how to filter the generated repertoire for individuals that transfer to the real world, as the transfer rate makes random sampling wasteful. Further work could address this by training a model to predict transferability from a limited real world dataset.

Our method allows for the generation of large-scale, diverse datasets for robotics, including for challenging tasks like grasping, without limitations about the applicable robots or policies. Instead, it requires defining relevant behavior spaces and the corresponding eligibility criteria. This is usually not a major issue, as it is often quite straightforward to define behavior components to promote diversity in a way that helps discover and explore the task solutions space.

The final test of NSMBS shall be the use of the generated repertoires as datasets to train closed loop policies \cite{levine,gan}, and the evaluation of those policies on real robots. The large size and high diversity of the generated datasets, the good sample efficiency of the method and its applicability to various problems and robots without adaptation or parameter tweaking, make it promising in this regard.

\section*{ACKNOWLEDGMENT}

This work was supported by ANR projects InDex and Learn2Grasp.


\bibliography{bib}
\bibliographystyle{bibst}

\end{document}